\documentclass[conference,letterpaper,10pt]{IEEEtran}
\IEEEoverridecommandlockouts

\usepackage[utf8]{inputenc}
\usepackage[T1]{fontenc}
\usepackage[english]{babel}
\usepackage{times}

\usepackage{microtype}
\usepackage{xcolor}
\usepackage{graphicx}
\usepackage{booktabs}
\usepackage{tabularx}
\usepackage{array}
\usepackage{amsmath,amssymb}
\usepackage{enumitem}
\usepackage{listings}
\usepackage{tikz}
\usepackage{csquotes}
\usepackage{hyperref}
\usepackage{xurl}
\usepackage{etoolbox}
\usepackage[backend=biber,style=ieee]{biblatex}

\appto{\bibsetup}{\raggedright}

\usetikzlibrary{arrows.meta,positioning,fit,calc,shapes.geometric,backgrounds}

\definecolor{axonblue}{HTML}{0B4F9C}
\definecolor{axongray}{HTML}{F2F4F8}
\definecolor{axonrust}{HTML}{B7410E}
\definecolor{axoncpp}{HTML}{00599C}
\definecolor{axonshm}{HTML}{2E7D32}
\definecolor{axonquic}{HTML}{6A1B9A}
\definecolor{axondark}{HTML}{061833}
\definecolor{axoncyan}{HTML}{13BFEA}
\definecolor{orcidgreen}{HTML}{A6CE39}

\hypersetup{
  colorlinks=true,
  linkcolor=axonblue,
  citecolor=axonblue,
  urlcolor=axonblue,
  pdftitle={AXON: A ROS 2 RMW with Shared-Memory/QUIC Transport and QKD/ML-KEM Key Establishment},
  pdfauthor={Sergio Sanchez de la Fuente, Miguel Angel Gonzalez-Santamarta, Francisco Javier Rodriguez-Lera, Vicente Matellan Olivera, Angel Manuel Guerrero-Higueras},
  pdfkeywords={ROS 2, RMW, shared memory, QUIC, QKD, ML-KEM}
}

\newcommand{\project}{\textsc{Axon}}

\makeatletter
\newcommand{\axon@brk}{\discretionary{}{}{}}
\begingroup
  \catcode`\_=\active \catcode`\-=\active \catcode`\/=\active
  \catcode`\.=\active \catcode`\:=\active
  \gdef\axon@literalchars{%
    \def_{\char`\_\axon@brk}%
    \def-{\char`\-\axon@brk}%
    \def/{\char`\/\axon@brk}%
    \def.{\char`\.\axon@brk}%
    \def:{\char`\:\axon@brk}%
  }
\endgroup
\newcommand{\axon@literal}[2]{%
  \begingroup
    \ttfamily#1%
    \catcode`\_=\active \catcode`\-=\active \catcode`\/=\active
    \catcode`\.=\active \catcode`\:=\active
    \axon@literalchars
    \endlinechar=-1\relax
    \scantokens{#2}%
  \endgroup}
\DeclareRobustCommand{\code}[1]{\axon@literal{}{#1}}
\DeclareRobustCommand{\env}[1]{\axon@literal{\slshape}{#1}}
\makeatother

\newcommand{\rmwaxon}{\code{rmw_axon}}
\newcommand{\axoncore}{\code{axon_core}}

\setlist[itemize]{topsep=2pt,itemsep=1pt,parsep=0pt,leftmargin=1.3em}
\setlist[enumerate]{topsep=2pt,itemsep=1pt,parsep=0pt,leftmargin=1.5em}

\newcolumntype{Y}{>{\raggedright\arraybackslash}X}

\lstdefinestyle{axoncode}{
  basicstyle=\ttfamily\scriptsize,
  breaklines=true,
  frame=single,
  framerule=0.3pt,
  backgroundcolor=\color{axongray},
  columns=fullflexible,
  keepspaces=true,
  showstringspaces=false,
  aboveskip=4pt,belowskip=4pt
}
\title{\LARGE \bf AXON: A ROS 2 RMW with Shared-Memory/QUIC Transport
and QKD/ML-KEM Key Establishment}

\newcommand{\orcid}[1]{\href{https://orcid.org/#1}{\textcolor{orcidgreen}{\raisebox{0.15ex}{\scriptsize$\bullet$}}}}

\author{
\IEEEauthorblockN{Sergio Sánchez de la Fuente\,\orcid{0009-0000-8748-8923}}
\IEEEauthorblockA{\textit{Robotics Group}\\
\textit{Universidad de León}\\
ssand@unileon.es}
\and
\IEEEauthorblockN{Miguel Ángel González-Santamarta\,\orcid{0000-0002-7658-8600}}
\IEEEauthorblockA{\textit{Robotics Group}\\
\textit{Universidad de León}\\
mgons@unileon.es}
\and
\IEEEauthorblockN{Francisco Javier Rodríguez-Lera\,\orcid{0000-0002-8400-7079}}
\IEEEauthorblockA{\textit{Robotics Group}\\
\textit{Universidad de León}\\
fjrodl@unileon.es}
\and[\hfill\mbox{}\hfill\par\hfill\mbox{}\hfill]
\IEEEauthorblockN{Vicente Matellán Olivera\,\orcid{0000-0001-7844-9658}}
\IEEEauthorblockA{\textit{Robotics Group}\\
\textit{Universidad de León}\\
vicente.matellan@unileon.es}
\and
\IEEEauthorblockN{Ángel Manuel Guerrero-Higueras\,\orcid{0000-0001-8277-0700}}
\IEEEauthorblockA{\textit{Robotics Group}\\
\textit{Universidad de León}\\
am.guerrero@unileon.es}
}

\begin{document}
\maketitle

\begin{abstract}
Robot Operating System~2 (ROS~2) standardizes application code against a
middleware interface (RMW) whose reference implementations are built on the Data
Distribution Service (DDS). We present \project{}, an alternative ROS~2 RMW
implementation that separates transport policy by deployment scope. A Rust core
and C++ adapter use POSIX shared-memory rings for same-host communication, QUIC
for remote communication, and a daemon for discovery and graph synchronization.
We then describe two fail-closed TLS~1.3 key-establishment configurations for
remote traffic. The \code{classic} configuration offers only the hybrid
X25519MLKEM768 group, preventing negotiation of a classical-only group. The
\code{qkd} configuration imports a 256-bit key obtained through the ETSI GS
QKD~014 API as a pairwise external PSK and offers no Diffie--Hellman group. Its
default \code{messages10} strategy additionally protects remote application
messages with AES-256-GCM, rotating KME material after ten outgoing messages
and using a fresh nonce per envelope; \code{session} relies on QUIC protection
alone. The external-PSK path requires a narrow extension to rustls, now bundled
with \project{}. We define the threat model, distinguish peer authentication in
the two configurations, and delimit the implementation-level validation from
ROS~2 conformance, comparative performance, and physical-QKD validation.
\end{abstract}


\section{Introduction}
ROS~2 decouples application code from the communication layer through the ROS
middleware interface (RMW)~\cite{ros2_middleware_interface,macenski2022ros2}. An
RMW implements a fixed C surface (\code{rmw_*}) that \code{rcl}, \code{rclcpp},
\code{rclpy}, command-line tooling, RViz, and Gazebo call, so any conforming
middleware can be selected at run time with
\env{RMW_IMPLEMENTATION}. The default implementations are built on the Data
Distribution Service (DDS) and its RTPS wire protocol~\cite{ros2_on_dds,
omg_ddsi_rtps_25}. DDS provides mature, standardized interoperability across
vendors, which is valuable when heterogeneous systems must interoperate on a
shared network.

That same universality is also a constraint. A DDS domain treats every peer
relationship as the same network problem, regardless of whether the two
endpoints share a host, a wired LAN, a lossy WiFi link, or a NAT boundary. In
practice a robotics system rarely looks uniform: a camera node and a perception
node on one workstation, a Gazebo simulation feeding a remote RViz over WiFi, and
a container talking to a physical robot each stress the transport differently.
The 2021 ROS Middleware Evaluation~\cite{ros_middleware_evaluation_2021} and the
emergence of non-DDS RMWs such as \code{rmw_zenoh}~\cite{rmw_zenoh_repo,
zenoh2023} show that the community increasingly treats transport policy as a
degree of freedom rather than a fixed cost.

\project{} takes that position deliberately. It implements the ROS~2 RMW API
with the QoS limitations stated in Section~\ref{sec:limits}, but \emph{does not
preserve DDS wire interoperability}. \project{} does not attempt to join an arbitrary
DDS/RTPS domain; it optimizes the path between ROS processes that intentionally
select it. This trade is explicit: users who need to interoperate with an
existing DDS deployment should keep a DDS RMW; users who control both endpoints
gain a transport tuned to how their system is actually deployed.

On top of that substrate, \project{} treats confidentiality as a first-class,
forward-looking concern. Robot fleets increasingly carry sensitive perception,
mapping, and command data over shared or wireless links, and their operational
lifetimes are long enough that ``harvest-now, decrypt-later'' adversaries and
future quantum computers are part of the threat
model~\cite{shor1994,mosca2018}. \project{} therefore makes the establishment of
the session key the object of design. Its default \code{classic} mode requires
a CRYSTALS-Kyber (ML-KEM) hybrid key
exchange~\cite{fips203,kyber2018} and offers nothing else, so a classical
fallback cannot be negotiated. Its \code{qkd} mode replaces the exchange
entirely with key material delivered by an ETSI-standard Quantum Key
Distribution management entity~\cite{etsi_qkd_014,bb84}, imported directly into
the TLS~1.3 key schedule. The optional \code{session} strategy leaves payload
protection to QUIC. The default \code{messages10} strategy deliberately adds a
second, application-level AEAD envelope so KME-delivered data keys can be
rotated independently of the QUIC connection.

\textbf{Contributions.} This paper makes the following
contributions:
\begin{enumerate}
  \item A two-layer ROS~2 RMW implementation (Rust core + C++ adapter) that routes
  same-host traffic over shared memory and cross-host traffic over QUIC, with a
  daemon-based control plane for discovery and graph introspection
  (Sections~\ref{sec:arch}--\ref{sec:ros}).
  \item A transport-security design in which QUIC protects remote ROS traffic
  and offers two mutually exclusive key-establishment modes: a mandatory hybrid
  ML-KEM exchange with no classical fallback, and quantum-distributed key
  material imported as a TLS~1.3 external pre-shared key. QKD deployments can
  additionally rotate application-layer KME keys every ten outgoing messages
  (Section~\ref{sec:security}).
  \item An implementation of the latter that required extending rustls, whose
  public API does not expose TLS~1.3 external PSKs --- the component of this
  work that configuration alone cannot reproduce
  (Section~\ref{sec:security}).
  \item Fail-closed enforcement throughout: mismatched peers are rejected via a
  mode field carried in discovery, obsolete configuration variables from earlier
  designs are refused at startup, and no path falls back to a classical-only
  exchange or to plaintext (Section~\ref{sec:security}).
  \item A validation scope covering ROS-facing behavior, local and remote
  transport paths, and fail-closed security behavior, with explicit limits on
  conformance, performance, and physical-QKD claims
  (Section~\ref{sec:eval}).
\end{enumerate}

\section{Background and Related Work}
\subsection{ROS~2, RMW, and DDS}
ROS~2 messages are described in IDL and serialized in Common Data Representation
(CDR). The RMW contract covers publishers, subscriptions, services, graph
introspection, wait sets, and QoS~\cite{ros2_qos_design}. DDS-based RMWs
(Fast~DDS~\cite{fastdds}, Cyclone~DDS~\cite{cyclonedds},
Connext~\cite{rti_connext}) implement this
contract over RTPS~\cite{omg_ddsi_rtps_25}. \code{rmw_zenoh} replaced RTPS with
the Zenoh protocol~\cite{zenoh2023} while keeping the RMW surface, demonstrating
that the RMW abstraction admits non-DDS transports. \project{} follows the same
substitutability principle but optimizes for controlled, self-selected
deployments and adds a native security path.

\subsection{Transports}
QUIC~\cite{rfc9000,rfc9001} provides multiplexed, encrypted and
connection-oriented streams over UDP, with modern loss recovery. That fits
cross-host robot links better than raw UDP multicast does for bulk sensor
data. \project{} uses the
\code{quinn}~\cite{quinn_docs} QUIC implementation. Same-host traffic instead
uses POSIX shared memory~\cite{linux_shm_overview} with
\code{eventfd}~\cite{linux_eventfd} notification and \code{epoll}~\cite{linux_epoll}
multiplexing, avoiding kernel copies for large frames.

\subsection{Quantum and post-quantum key agreement}
QKD distributes symmetric key material whose secrecy rests on physics rather than
computational hardness~\cite{bb84,pirandola2020advances}; the ETSI GS QKD~014
specification standardizes a REST API through which applications request keys
from a Key Management Entity (KME)~\cite{etsi_qkd_014}. QKD requires specialized
hardware or a simulator such as QuKayDee~\cite{qukaydee}; finite key supply and
KME request latency require an explicit consumption policy. Post-quantum cryptography instead
resists quantum attacks computationally; ML-KEM (CRYSTALS-Kyber),
standardized in FIPS~203~\cite{fips203,kyber2018}, is a lattice-based key
encapsulation mechanism. The current TLS working-group specification defines
hybrid groups that combine an ephemeral classical exchange with ML-KEM,
including \code{X25519MLKEM768}~\cite{ietf_ecdhe_mlkem}. \project{} offers only
that group in its classic configuration. Its QKD configuration uses TLS~1.3
external pre-shared keys~\cite{rfc8446,rfc9257}, which allow high-entropy key
material provisioned out of band to seed the TLS key schedule. The pairwise
derivation applies project-specific domain separation and includes the KME
\code{key_ID} in both the HKDF context and the TLS identity; the exact
construction, its relation to external-PSK importers~\cite{rfc9258}, and its
limitations are stated in Section~\ref{sec:security}.

\section{System Architecture}
\label{sec:arch}
\begin{figure}[t]
\centering
\resizebox{\columnwidth}{!}{%
\begin{tikzpicture}[
  font=\scriptsize,
  box/.style={rounded corners=2pt,draw,align=center,minimum height=6mm,inner sep=3pt},
  lay/.style={rounded corners=3pt,draw,thick,inner sep=5pt}
]
\node[box,fill=axoncpp!5,draw=axoncpp,densely dashed,text=axoncpp,
      minimum width=58mm] (stack)
  {ROS~2 stack: \code{rcl} / \code{rclcpp} / \code{rclpy} / RViz / Gazebo};
\node[below=4mm of stack,box,fill=axoncpp!12,draw=axoncpp,minimum width=58mm]
  (rmw) {\rmwaxon{} (C++)\\\code{rmw_*} API, FastCDR};
\node[below=4mm of rmw,box,fill=axongray,draw=black,minimum width=58mm]
  (ffi) {\code{extern "C"} FFI \; (\code{axon_session_*})};
\node[below=9mm of ffi,box,fill=axonshm!14,draw=axonshm,minimum width=24mm,
      anchor=north east,xshift=-1.5mm] (shm) at (ffi.south)
  {SHM rings\\+ \code{eventfd}};
\node[box,fill=axonquic!14,draw=axonquic,minimum width=24mm,
      anchor=north west] (quic) at ([xshift=3mm]shm.north east)
  {QUIC\\(\code{quinn})};
\node[below=3mm of shm,box,fill=axongray,draw=black,minimum width=24mm]
  (sess) {session /\\graph / QoS};
\node[below=3mm of quic,box,fill=axongray,draw=black,minimum width=24mm]
  (sec) {security\\QKD + Kyber};
\begin{scope}[on background layer]
  \node[lay,fill=axonrust!8,draw=axonrust,fit=(shm)(quic)(sess)(sec)] (core) {};
\end{scope}
\node[text=axonrust,font=\scriptsize\bfseries,anchor=south west,inner sep=0pt]
  at ([yshift=1.2mm]core.north west) {\axoncore{} (Rust)};
\node[box,fill=axoncyan!18,draw=axonblue,minimum height=13mm,text width=17mm,
      anchor=west] (daemon) at ([xshift=6mm]core.east |- quic)
  {\code{axon_daemon}\\UDP discovery};
\draw[-{Latex}] (stack) -- (rmw);
\draw[-{Latex}] (rmw) -- (ffi);
\draw[-{Latex}] (ffi.south) -- (core.north);
\draw[<->,axonblue] (quic.east) -- (daemon.west);
\end{tikzpicture}
}
\caption{\project's two-layer architecture. A thin C++ adapter implements the
ROS~2 \code{rmw_*} contract and calls a Rust core through a stable C FFI. The
core owns shared-memory (local) and QUIC (remote) data planes; an auto-spawned
daemon owns discovery and graph synchronization.}
\label{fig:arch}
\end{figure}
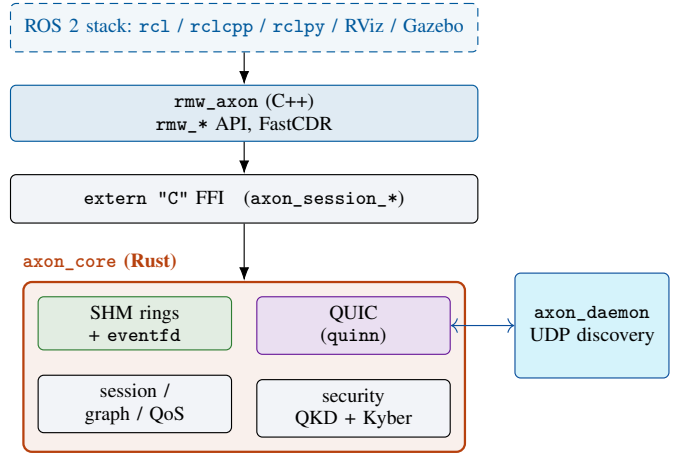

\project{} is split into two layers connected by a stable \code{extern "C"} FFI
boundary (Fig.~\ref{fig:arch}).

\textbf{C++ RMW adapter (\rmwaxon).} Implements the RMW entry points used by
the supported ROS~2 scenarios, handles ROS type support and CDR encapsulation via
FastCDR~\cite{fastcdr_docs}, and forwards all transport decisions to the core.
It contains no networking logic of its own.

\textbf{Rust core (\axoncore).} Owns all transport and coordination: SHM ring
buffers, the discovery daemon client, QUIC data transport, session lifecycle and
routing, the graph cache for introspection, the \code{epoll} wait set, QoS
interpretation, ACL and encryption, CDR serialization, subscriber queues, and
leaky-bucket rate limiting. Rust's ownership and type systems prevent several
classes of memory-safety defect in safe code, although the C FFI boundary still
requires explicit review.

\textbf{Session model.} Each \code{rmw_init}/\code{rmw_shutdown} pair is an
isolated \emph{session} with its own transport endpoints and graph cache. Local
SHM rings and their \code{eventfd} rendezvous sockets are keyed by
\env{ROS_DOMAIN_ID}, so stale samples cannot leak between domains.

\textbf{Control plane vs.\ data plane.} \project{} separates the two
explicitly. The control plane (daemon discovery, graph synchronization, security
negotiation) establishes \emph{that} a peer and route exist; the data plane (SHM
or QUIC) moves samples. Many middleware failures become tractable once these are
distinguished: a graph can be correct while a route is unreachable, and a route
can be healthy while a wait set fails to wake.

\section{Local Transport: Shared Memory}
\label{sec:shm}
For same-host peers, copying large sensor frames through the network stack is
wasteful. \project{} allocates a POSIX shared-memory
object~\cite{linux_shm_overview} per topic and maps it into publisher and
subscriber processes as a ring of fixed-size slots. A publisher writes into the
next slot and signals readers through an \code{eventfd}~\cite{linux_eventfd};
readers are woken through the \code{epoll}-based wait set~\cite{linux_epoll}. The
eventfds themselves are passed between processes over Unix-domain sockets using
\code{SCM_RIGHTS}.

Ring depth is sized from QoS and message type. High-rate bursty topics such as a
Gazebo \code{/clock} keep a burst-tolerant minimum depth even when the ROS QoS
depth is one, so a slow consumer recovers to the newest retained sample rather
than failing a take. The per-ring data budget is bounded
(\env{AXON_RING_BUFFER_SIZE_MB}) and clamps slot counts for very large message
types (images, point clouds). Shared memory is a host resource protected by
filesystem permissions; \project{} treats a same-host peer with sufficient OS
privilege as trusted, and applies encryption only to traffic that leaves the
host (Section~\ref{sec:security}).

\section{Discovery and the Daemon}
\label{sec:daemon}
\project{} auto-spawns an \code{axon_daemon} that owns discovery and graph
exchange. It uses UDP multicast (with a broadcast fallback on every non-loopback
IPv4 interface) to find peers, and shared memory to exchange local
graph/match information with co-located sessions. A daemon-backed \code{GraphCache}
answers topic, service, node, publisher, and subscription queries, so standard
\code{ros2 topic/node/service} tooling works. Static peers can be configured with
\env{AXON_DAEMON_PEERS} for networks where multicast is blocked, and advertised
addresses can be overridden (\env{AXON_ADVERTISE_ADDRS}) so a container behind a
NAT bridge does not advertise an unreachable address. Because discovery lives in
a separate process, a data-plane fault does not corrupt the graph, and a fresh
session can re-synchronize from the daemon's warm state.

\section{Remote Transport: QUIC}
\label{sec:quic}
Once the daemon has discovered a remote peer, cross-host data flows over
QUIC~\cite{rfc9000,rfc9001} using \code{quinn}~\cite{quinn_docs}. QUIC gives
\project{} multiplexed streams, removing transport-layer head-of-line blocking
between streams; streams still share connection congestion control and sender
resources. Defaults target local robot networks: an MTU-safe 1200-byte UDP
payload avoids IP fragmentation on WiFi, a 16~MiB connection send window leaves
room for one large sensor frame alongside small control streams, and remote
payloads above 1~KiB are zstd-compressed by
default~\cite{rfc8878}. The remote send pipeline is ordered per (peer, topic):
reliable publishers use a lossless sender queue, whereas best-effort publishers
retain at most their configured history depth under backpressure. Services map
to two internal request/response topics with type-bounded allocation, and route
over the same QUIC path when the peer is remote.

\section{QoS, Graph, Services, and Actions}
\label{sec:ros}
\project{} implements a documented subset of ROS~2 QoS behavior and checks
publisher/subscription
compatibility: reliability (best-effort/reliable), durability
(volatile/transient-local), history (\code{KEEP_LAST}/\code{KEEP_ALL}), and
deadline, lifespan, and liveliness. Graph introspection is treated as a
correctness property, not a convenience: reporting a wrong type or endpoint count
is a semantic bug, because tools and users act on that metadata. Services are
implemented as paired internal request/response topics with response ownership
tracked per client, so multiple clients of one service receive their own replies.
Actions build on services and topics. Wait sets multiplex subscription, service,
and guard-condition eventfds through \code{epoll} with optional timerfd polling,
with a shorter poll slice when services/clients are present to avoid false
lifecycle timeouts under load. Remote reliability has not yet been qualified as
DDS-equivalent; its resource trade-offs and bounded receive histories are
stated in Section~\ref{sec:limits}.

\section{Transport Security Design}
\label{sec:security}
\project{} always uses QUIC packet protection for remote ROS traffic. In QKD
\code{messages10} mode it additionally applies an application-level AEAD
envelope to remote topic, service, and action payloads; QKD \code{session} and
\code{classic} add no such envelope. UDP discovery announcements remain
unauthenticated and unencrypted. Same-host shared-memory payloads never enter
QUIC and are not encrypted; their boundary is the Unix account,
\code{/dev/shm} permissions, and host integrity.

What \project{} does contribute is \emph{how the key that protects that path is
established}. Two mutually exclusive modes are selected with
\env{AXON_SECURITY_MODE} (Table~\ref{tab:modes}), and both are fail-closed:
neither can negotiate its way down to a weaker construction.

\begin{table}[t]
\caption{The two \project{} security modes. Both use QUIC 1-RTT packet
protection, but differ in key establishment, peer authentication, and forward
secrecy.}
\label{tab:modes}
\centering\small
\begin{tabularx}{\columnwidth}{@{}lYY@{}}
\toprule
& \textbf{\code{classic}} & \textbf{\code{qkd}} \\
\midrule
TLS establishment & Hybrid X25519 + ML-KEM-768 & ETSI QKD 014 key as external PSK \\
Key-exchange groups & Exactly one & \emph{Empty} \\
Application AEAD & None & Optional; default rotates every 10 messages \\
Forward secrecy & Yes (ephemeral DH) & No (\code{psk_ke}) \\
Peer authentication & No (current certificate policy) & Pairwise PSK \\
Fallback & None & None \\
\bottomrule
\end{tabularx}
\end{table}

\subsection{Threat model and authentication boundary}
The remote channel is intended to resist passive capture, packet modification,
and replay after the TLS handshake has authenticated the relevant channel
secret. The two configurations do not provide identical peer authentication.
In \code{qkd}, a high-entropy pairwise PSK authenticates possession of the
KME-delivered secret, subject to correct KME provisioning and identity
binding~\cite{rfc9001,rfc9257}. In the current \code{classic} configuration,
certificates are self-signed and accepted without a CA or pinned fingerprint.
The hybrid exchange therefore protects recorded traffic against later
cryptanalysis, but it does not authenticate a peer against an active
man-in-the-middle. Deployments that require active-peer authentication must add
certificate validation or pinning; this is future work.
The QKD authentication statement likewise assumes that each SAE credential is
provisioned exclusively to its intended host. The bundled QuKayDee beta
profiles trade that property for deployment convenience and therefore are test
fixtures, not production identity provisioning.

\subsection{Design goals}
\begin{itemize}
  \item \textbf{Explicit protection layers.} QUIC/TLS always protects remote
  traffic; \code{messages10} optionally adds KME-keyed application AEAD.
  \item \textbf{No silent downgrade.} If the configured mode cannot be
  established, communication does not happen.
  \item \textbf{Bounded key consumption.} \code{session} consumes one KME key
  per daemon pair, while \code{messages10} amortizes each data key across ten
  outgoing messages.
  \item \textbf{Standard primitives only.} \project{} contributes configuration
  and integration, not new cryptography.
\end{itemize}

\subsection{Classic mode: mandatory hybrid key exchange}
In \code{classic} mode the rustls crypto provider is built with a
key-exchange group list containing exactly one entry, \code{X25519MLKEM768},
and the protocol version is pinned to TLS~1.3. That group is itself hybrid: the
derived secret depends on both the classical X25519 exchange and ML-KEM-768
(CRYSTALS-Kyber, FIPS~203)~\cite{fips203,kyber2018}. Under the assumptions of
the hybrid specification, the combiner is intended to retain secrecy while at
least one component remains secure~\cite{ietf_ecdhe_mlkem}.

Because the list has a single member, there is no classical-only group to
negotiate down to, and a peer that cannot perform the hybrid exchange is
rejected rather than accommodated. This is a deliberate departure from the
rustls provider's multi-group defaults. The \code{prefer-post-quantum} feature
changes priority but does not by itself remove classical groups
from the provider~\cite{rustls_defaults}.

\subsection{QKD mode: quantum key material inside TLS}
In \code{qkd} mode the key does not come from a Diffie-Hellman exchange at all.
The daemon obtains a 256-bit key from its local Key Management Entity following
ETSI GS QKD~014~\cite{etsi_qkd_014}: the daemon with the lower identifier calls
\code{enc_keys}, and the peer retrieves the matching value by its public
\code{key_ID} with \code{dec_keys}. Only that identifier travels over
\project's discovery channel; the key itself never does.

\project{} then derives a TLS-specific 256-bit secret with
HKDF-SHA256~\cite{rfc5869} under the domain label
\code{AXON-QKD-TLS13-EXTERNAL-PSK-v1} and imports it into the TLS~1.3 key
schedule as an \emph{external pre-shared key}, together with a namespaced
identity containing the KME \code{key_ID}. The deployment assumes that each
KME key is restricted to one SAE/daemon pair. This is a project-specific
domain-separated derivation, not a claim of conformance to the RFC~9258
importer. The client offers
\code{psk_ke} with no
\code{supported_groups} and no \code{key_share}; the QKD provider's
key-exchange group list is empty. The server resolves the identity, verifies the
TLS binder, and only then selects the PSK.

The bootstrap KME key protects the session through the standard TLS key
schedule. A successful handshake proves that both endpoints hold the same
pairwise PSK. In a physical QKD deployment that PSK originates in
quantum-distributed material; a simulator exercises only the API integration.
The handshake cannot fall back to X25519, to ML-KEM, to a certificate-only
exchange, or to plaintext, because none of those paths is configured.

\textbf{Application-key strategies.} The environment variable
\env{AXON_QKD_KEY_MODE} selects one of two strategies. \code{session} uses the
bootstrap PSK only and leaves application protection to QUIC. The default
\code{messages10} strategy wraps each remote topic or service message in an
\code{AXQKD001} envelope before it enters QUIC. The sender obtains a 256-bit
KME key for the receiving SAE and uses it for at most ten outgoing messages,
with a fresh 96-bit nonce for every AES-256-GCM encryption. The authenticated
metadata binds the message kind, topic hash, sender SAE, and KME \code{key_ID}.
The receiver retrieves the matching key with \code{dec_keys}; co-located ROS
processes coordinate that retrieval through a user-only shared-memory cache.
The SAE and \code{key_ID} travel on the wire, but the key bytes do not. Topic
fan-out to several processes on one remote host reuses one sealed copy of the
same logical publication rather than consuming a distinct key use per process.

\textbf{Extending the TLS implementation.} This required work below
\project. rustls does not expose TLS~1.3 external PSKs through its public API,
so \project{} includes the narrow \code{rustls-axon}
fork~\cite{rustls_axon_repo}, pinned to an immutable commit of the official
release and vendored as a Git subtree.
The fork adds an external-PSK type with a public identity and
a zeroized secret, a server-side resolver, external-binder derivation using the
RFC~8446 \code{ext binder} label, and PSK-only handshake paths that carry no
key share. All record and packet protection remains upstream rustls and quinn
code. This extension cannot be reproduced through configuration alone.

\textbf{Security trade-offs.} \code{psk_ke} provides no independent Diffie-Hellman
forward secrecy. QKD-mode security rests on the bootstrap PSK remaining secret
and being erased when the daemon session ends; compromise of that key exposes
captured QUIC traffic from the same session. In \code{messages10}, compromise of
an application data key additionally exposes its batch of at most ten outgoing
messages. Classic mode retains ephemeral hybrid
key establishment and therefore forward secrecy against later key compromise,
but, as noted above, its current certificate policy does not prevent active
MITM. RFC~9257 recommends combining external PSKs with an ephemeral exchange
where possible; \project{} deliberately uses \code{psk_ke} to make QKD-key
possession the only configured key-establishment path~\cite{rfc9257}.

\subsection{Traffic protection}
Both modes use ordinary QUIC 1-RTT packet protection. QKD \code{messages10}
adds the application AEAD described above; its AES-256-GCM choice is independent
of the negotiated QUIC suite. The QUIC traffic cipher is
selected with \env{AXON_QUIC_CIPHER}: either ChaCha20-Poly1305~\cite{rfc8439},
the portable default, or AES-256-GCM. This paper does not claim a measured
performance advantage for either suite on the evaluated hosts. AES-128 is not
offered as a 1-RTT traffic suite.

One protocol constraint deserves explicit mention because it is easily
misread as a weakness: QUIC~v1 mandates AES-128-GCM for \emph{Initial} packet
protection. Those keys are derived from a published salt and the connection ID,
and are independent of the negotiated 1-RTT suite. Because an observer can
derive them, Initial packets are not considered to have confidentiality or
integrity protection~\cite{rfc9001}. No QUIC implementation can substitute a
different cipher there while remaining QUIC~v1 compliant.

\subsection{Fail-closed enforcement}
Configuration is validated before the daemon starts, so an invalid setup
surfaces as a startup error rather than as an RMW process waiting on a daemon
that correctly refuses to run. Startup rejects any mode other than
\code{classic} or \code{qkd}, any cipher other than the two above, QKD mode
without a complete KME and SAE configuration, and --- notably --- the
\emph{obsolete} variables from earlier designs (\env{AXON_SECURITY_PROFILE},
\env{AXON_ENCRYPTION_MODE}, \env{AXON_ENCRYPTION_KEY}), so that a stale
deployment script cannot silently produce a weaker configuration than its author
intended.

Connection setup rejects peers advertising a different security mode (the mode
travels in the daemon HELLO), classic peers that do not negotiate the hybrid
group, QKD peers with no locally available daemon-pair key, and unknown
external-PSK identities or incorrect binders.

\subsection{What this design does and does not claim}
After a valid handshake, QUIC packet protection provides confidentiality,
integrity, and replay protection for remote ROS payloads relative to the
authenticated channel secret~\cite{rfc9001}. In \code{qkd}, the pairwise PSK
also authenticates possession of the KME-delivered secret. In \code{classic},
these channel properties do not establish peer identity under the current
self-signed-certificate policy.

Accordingly, \project{} does not claim to be ``quantum-proof'' or
information-theoretically secure: QKD material seeds symmetric AEAD
keys and is not consumed as a one-time pad. \code{session} has constant
key-material cost per daemon pair; \code{messages10} consumes one application
key per batch of ten outgoing messages. UDP discovery is neither
authenticated nor confidential, so an active attacker can spoof announcements,
cause failed connections, observe traffic metadata, or deny service. Without
the pairwise PSK it cannot complete a \code{qkd} handshake, but it can mount a
MITM against \code{classic} because certificates are not validated against a
trust root. There are no per-node authenticated identities, certificate
revocation, or signed discovery. Local shared memory and a compromised endpoint
lie outside the remote cryptographic boundary. The KME link is a separate trust
boundary whose own post-quantum properties \project{} cannot upgrade. A QKD
\emph{simulator} validates the API integration but is not evidence of a physical
QKD link. Finally, \project{} has not undergone a third-party cryptographic
audit, and its QKD mode depends on a forked TLS implementation that must be
rebased and re-reviewed whenever the upstream version changes.

\section{Implementation}
\label{sec:impl}
\axoncore{} is a Rust crate (\textasciitilde{}25k source lines) exposing a multi-session C
FFI (\code{axon_session_create}, graph, and service APIs). \rmwaxon{} is a C++
shared library (\textasciitilde{}5k source/header lines) built by CMake, which first compiles
the Rust core (\code{cargo build --release}) and links the resulting static
library. QUIC is provided by \code{quinn} over \code{rustls} with the
\code{aws-lc-rs} cryptographic provider. Classic mode overrides the provider's
key-exchange group list with the single entry \code{X25519MLKEM768}; QKD mode
leaves that list empty and supplies an external PSK instead. The KME client uses
\code{reqwest} with mutual TLS, and key material is zeroized on drop.
Configuration validation runs before the daemon is spawned, so an invalid
fail-closed setup surfaces as a startup error rather than a hang.

QKD mode uses a fork of \code{rustls}~0.23.40, pinned to an immutable commit and
vendored under \code{third_party/rustls-axon}. It exposes TLS~1.3 external PSKs: an external-PSK type
with a public identity and zeroized secret, a server-side resolver,
external-binder derivation with the RFC~8446 \code{ext binder} label, and
PSK-only handshake paths carrying no key share. Record and packet protection
remain upstream code. This is an operational commitment as much as a technical
one: the fork must be rebased and re-reviewed on every upstream version change.
A normal clone contains the complete fork and requires no second repository or
SSH credential at build time.

The build produces a standard ROS~2 package selectable with
\env{RMW_IMPLEMENTATION}=\code{rmw_axon} across Humble through Rolling. Because
\project{} contributes configuration and protocol integration rather than new
primitives, its security claims reduce to the correctness of the underlying
libraries, the narrow forked API, the deployment's authentication policy, and
the enforcement paths we test.

An ordinary \code{colcon build} selects no SAE identity and leaves
\code{classic} as the runtime default. For reproducible two-host simulator
tests, the build option \code{AXON_QKD_ROLE=1} or \code{2} installs the
corresponding bundled QuKayDee beta profile beside the daemon. The resulting
installation can switch between \code{qkd} and \code{classic} at runtime
without separate KME-path variables; custom deployments instead provide their
own SAE material explicitly.

\textbf{Software availability.} The project source and build instructions are
maintained in the \href{https://github.com/AXON-rmw/rmw_axon}{AXON GitHub
repository}~\cite{rmw_axon_repo}. The TLS modifications are also maintained in
the \href{https://github.com/ssancd03/rustls-axon}{rustls-axon GitHub
repository}~\cite{rustls_axon_repo} and are fully vendored in the main tree. At
the time of writing, the linked repositories still require GitHub authorization;
a public main-tree release or archival snapshot is therefore required for
independent reproduction.

\section{Validation Scope}
\label{sec:eval}
The implementation is checked at three boundaries that correspond to its
architecture. Core checks exercise shared-memory and QUIC transport semantics,
including ordered remote delivery and the distinct backpressure policies for
reliable and best-effort publishers. ROS-facing checks cover publication and
subscription, request--response correlation for services and actions, graph
synchronization, wait-set readiness, and daemon restart behavior. Security
checks cover strict mode and cipher selection, the single hybrid group offered
by \code{classic}, the external-PSK-only configuration used by \code{qkd}, KME
request/response handling, fail-closed peer negotiation, and the authenticated
application envelope used by \code{messages10}.

This scope establishes implementation-level consistency for the mechanisms
described in this paper. It does not constitute a cryptographic proof, a
third-party security audit, qualification against the complete ROS~2 RMW
conformance suite, a comparative performance benchmark, or validation against a
physical QKD link. Those forms of evidence require separate controlled studies
and are not claimed here.

\section{Limitations}
\label{sec:limits}
\project{} is not DDS wire-compatible and cannot interoperate with an existing
DDS domain --- by design. Best-effort remote queues discard their oldest pending
sample at the configured history depth under backpressure. Reliable remote
queues preserve send order without a fixed queue bound, which avoids silent
sender-side loss but can increase memory use when a peer is persistently slow;
receive and shared-memory histories remain bounded. These semantics have not
been qualified as DDS-equivalent reliability. The implementation has
not been qualified with the complete ROS~2 RMW conformance suite. Certificates
are self-signed; \code{classic} therefore lacks authenticated peer identity,
whereas \code{qkd} authenticates pairwise PSK possession. QUIC congestion
control has not been tuned for shared WAN links. Some QoS events are observable
only by polling. The local data plane depends on Linux primitives.

Three limitations are specific to the security design. First, \code{qkd} mode
uses \code{psk_ke} and therefore has no independent Diffie-Hellman forward
secrecy: compromise of a session key exposes captured traffic from that session,
whereas \code{classic} mode retains ephemeral hybrid key establishment. Second,
the mode must match on both peers and mismatches are rejected, so enabling a mode
is a fleet-wide decision rather than a per-node one --- there is no gradual
migration path. The QKD application-key strategy must also match; unlike the
security mode, it is not advertised during discovery, so a mismatch manifests
as a data-path failure. Third, \code{qkd} mode depends on a forked TLS
implementation that must be rebased and re-reviewed whenever the upstream
version changes, although the pinned snapshot is bundled with \project{}.
The bundled QuKayDee identities are beta test fixtures: distributing their
private credentials removes exclusive SAE identity and is unsuitable for
production authentication claims.

Finally, we have not quantified either mode's handshake and steady-state
overhead, the redesigned security path has not yet been validated in a
controlled two-host campaign, and \project{} has not had a third-party
cryptographic audit. A public release or archival snapshot is still required
for independent reproduction.

\section{Conclusion}
\project{} demonstrates an alternative ROS~2 RMW architecture using shared
memory on the host, QUIC between hosts, and a daemon-based control plane, with
the API and QoS limitations documented above. On that substrate it makes key
establishment and rotation explicit: QUIC protects all remote traffic, and
\project{} constrains how the TLS key schedule is seeded --- either by a hybrid
X25519/ML-KEM exchange with no classical-only group, or by KME-delivered
material imported as an external PSK. The default QKD strategy additionally
rotates AES-256-GCM application keys after ten outgoing messages.

The second path required extending rustls because its public API does not expose
TLS~1.3 external PSKs. Other production TLS stacks do expose external-PSK
interfaces~\cite{rfc9257}; the claim is specific to the selected implementation,
not to TLS libraries in general. The post-quantum exchange, by contrast, is
configured through the rustls cryptographic provider.

Future work includes authenticated peer identity for \code{classic}, full RMW
conformance testing, congestion control for shared links, quantifying the
overhead of each mode, a two-host campaign against the redesigned path,
reproducible comparisons with DDS and Zenoh RMWs, third-party review of the
rustls fork, and validation against physical QKD hardware rather than a
simulator.

\printbibliography

\end{document}